\documentclass[sigconf,nonacm]{acmart}

\usepackage{algorithm}
\usepackage{algorithmic}
\usepackage{balance} 
\usepackage{xcolor}

\AtBeginDocument{%
  }
\usepackage{tcolorbox}
\tcbuselibrary{listings}
\tcbset{
  mypromptbox/.style={
    colback=gray!5,
    colframe=gray!60,
    fonttitle=\bfseries,
    title=Knowledge-enhanced SMILES Repair Prompt,
    listing only,
    listing options={
      basicstyle=\ttfamily\small,
      breaklines=true,
    },
    boxrule=0.4pt,
    sharp corners,
    coltitle=black,
  }
}
\tcbset{
  mypromptbox/.style={
    colback=gray!5,
    colframe=gray!60,
    fonttitle=\bfseries,
    listing only,
    listing options={
      basicstyle=\ttfamily\small,
      breaklines=true,
    },
    boxrule=0.4pt,
    sharp corners,
    coltitle=black,
  }
}

\begin{document}

\title{Empowering LLMs for Structure-Based Drug Design via Exploration-Augmented Latent Inference}


\author{Xuanning Hu}
\orcid{0009-0006-4373-0670}
\affiliation{%
  \institution{College of Computer Science and Technology, Jilin
University}
  \institution{Key Laboratory of Symbolic Computation and Knowledge
Engineering of Ministry of Education, Jilin University}
  \city{Changchun}
  \state{Jilin}
  \country{China}}
\email{huxn24@mails.jlu.edu.cn}

\author{Anchen Li}
\orcid{0000-0001-9828-6964}
\affiliation{%
  \institution{Department of Computer Science, Aalto University}
  \city{Espoo}
  \country{Finland}}
\email{anchen.li@aalto.fi}

\author{Qianli Xing}
\orcid{0000-0002-3224-0928}
\authornote{Corresponding author.}
\affiliation{%
  \institution{College of Computer Science and Technology, Jilin
University}
  \institution{Key Laboratory of Symbolic Computation and Knowledge
Engineering of Ministry of Education, Jilin University}
  \city{Changchun}
  \state{Jilin}
  \country{China}}
\email{qianlixing@jlu.edu.cn}

\author{Jinglong Ji}
\orcid{0009-0009-6971-7636}
\affiliation{%
  \institution{College of Artificial Intelligence, Jilin
University}
  \institution{Key Laboratory of Symbolic Computation and Knowledge
Engineering of Ministry of Education, Jilin University}
  \city{Changchun}
  \state{Jilin}
  \country{China}}
\email{jijl22@mails.jlu.edu.cn}

\author{Hao Tuo}
\orcid{0009-0005-3612-582X}
\affiliation{%
  \institution{College of Computer Science and Technology, Jilin
University}
  \institution{Key Laboratory of Symbolic Computation and Knowledge
Engineering of Ministry of Education, Jilin University}
  \city{Changchun}
  \state{Jilin}
  \country{China}}
\email{tuohao25@mails.jlu.edu.cn}

\author{Bo Yang}
\orcid{0000-0002-5559-2547}
\authornotemark[1]
\affiliation{%
  \institution{College of Computer Science and Technology, Jilin
University}
  \institution{Key Laboratory of Symbolic Computation and Knowledge
Engineering of Ministry of Education, Jilin University}
  \city{Changchun}
  \state{Jilin}
  \country{China}}
\email{ybo@jlu.edu.cn}

\renewcommand{\shortauthors}{Xuanning Hu et al.}

\begin{abstract}
Large Language Models (LLMs) possess strong representation and reasoning capabilities, but their application to structure-based drug design (SBDD) is limited by insufficient understanding of protein structures and unpredictable molecular generation. To address these challenges, we propose Exploration-Augmented Latent Inference for LLMs (ELILLM), a framework that reinterprets the LLM generation process as an encoding, latent space exploration, and decoding workflow. ELILLM explicitly explores portions of the design problem beyond the model’s current knowledge while using a decoding module to handle familiar regions, generating chemically valid and synthetically reasonable molecules. In our implementation, Bayesian optimization guides the systematic exploration of latent embeddings, and a position-aware surrogate model efficiently predicts binding affinity distributions to inform the search. Knowledge-guided decoding further reduces randomness and effectively imposes chemical validity constraints. We demonstrate ELILLM on the CrossDocked2020 benchmark, showing strong controlled exploration and high binding affinity scores compared with seven baseline methods. These results demonstrate that ELILLM can effectively enhance LLMs' capabilities for SBDD. Our code is available at https://github.com/hxnhxn/ELILLM.
\end{abstract}




\maketitle

\section{Introduction}
Structure-Based Drug Design (SBDD), which aims to design novel compounds conditioned on structural information specific to the protein binding site, is a fundamental problem in drug discovery and biomedical research, and has been widely applied in tasks such as virtual screening and lead compound identification. Early approaches adopted variational autoencoders (VAEs)~\cite{rf:vae20,rf:ligan}, autoregressive models~\cite{rf:ar,rf:graphbp,rf:pocket2mol}, and more recently, diffusion models~\cite{rf:targetdiff,rf:decompdiff,rf:ipdiff,rf:alidiff}—all of which have shown strong performance in generating valid and diverse molecules. However, unlike human experts who iteratively refine molecular designs based on accumulated feedback, these models typically follow a generation process that lacks intermediate feedback and relies heavily on the training data distribution, which makes their generation less flexible and prevents them from dynamically incorporating new experience.

Large Language Models (LLMs), with their strong generalization and reasoning abilities, offer a promising way to address these limitations. Equipped with good understanding of natural language, strong representation skills, and flexible generation abilities, LLMs are suitable for step-by-step molecular design guided by intermediate feedback. This allows more intuitive interactions and the possibility to follow expert-like, iterative design processes, making AI-assisted drug discovery easier to use. Their success in complex reasoning tasks such as code generation~\cite{rf:llmfunsearch}, theorem proving~\cite{rf:llmprove}, and strategic planning~\cite{rf:llmplan} shows that these abilities can also be applied to more interactive molecular design.

However, LLMs face significant challenges in SBDD. \citet{rf:llmprotein} point out that LLMs do not have enough understanding of protein sequences, likely because their pretraining data does not include enough relevant information. \citet{rf:llmsober} report that LLMs are less accurate than traditional models in predicting molecular properties. These results suggest that current LLMs are unlikely to generate target molecules reliably. \citet{rf:lmlf} tried to guide LLMs for targeted generation using logical feedback, but the generated molecules were often similar to the input compounds and appeared almost random, showing weak targeting performance. To address this problem, some studies avoid using general LLMs and instead train specialized chemical GPT models on large protein–ligand datasets~\cite{rf:cmolgpt,rf:tokenmol,ref:tamgen}. However, this approach reduces the natural language capabilities of LLMs and still faces the same limitations as traditional deep learning models, being restricted by the distribution of training data. Therefore, finding effective ways to guide LLMs for SBDD remains an important and meaningful research direction.

To this end, we propose \textbf{Exploration-Augmented Latent Inference for Large Language Models (ELILLM)}, a framework that reinterprets the LLM generation process as an encoding, exploration, and decoding workflow for SBDD. ELILLM leverages the latent space of LLMs to explicitly explore portions of the design problem that fall outside the model’s current understanding, such as generating molecules with optimized binding to the target protein, while allowing the model to handle familiar regions effectively through the decoding module, producing chemically valid and synthetically reasonable molecules. Specifically, Bayesian optimization is applied in the latent space to systematically explore latent embeddings, enhancing the LLM’s ability to generate molecules aligned with the structural information of the target protein. To handle variable-length embeddings in the latent space, we introduce a \textit{position-aware surrogate model}. By using a position-invariant aggregation strategy, this model efficiently predicts the affinity distribution corresponding to latent embeddings, guiding exploration in the latent space. Finally, we develop a \textit{knowledge-guided LLM decoding} strategy that repairs candidate embeddings using chemical knowledge, reducing the randomness in LLM decoding and producing ligand molecules that satisfy task-specific requirements.
Our main contributions are summarized as follows.
\begin{itemize}
\item We propose a novel framework, \textbf{ELILLM}, that enhances LLMs for structure-based drug design by explicitly exploring the latent space. The framework reinterprets the LLM generation process as an encoding, exploration, and decoding workflow, allowing targeted exploration of regions outside the LLM’s current understanding while leveraging its strengths for familiar regions. The workflow is clearly defined, making it easy to apply ELILLM to SBDD tasks.

\item We implement this framework for SBDD using a Bayesian optimization-based exploration strategy. The approach consists of three main processes: Molecular-Guided LLM Encoding, BO-Based Latent Space Exploration, and Knowledge-Guided LLM Decoding. This implementation demonstrates how ELILLM can efficiently simulate expert-like iterative molecular design with interpretability and control.

\item We perform a comprehensive evaluation on the CrossDocked2020 benchmark, ensuring consistent Vina docking scores for identical SMILES. The results show that ELILLM improves LLMs’ ability to perform controlled exploration in the latent space, generating molecules with higher predicted binding affinity. Our method achieves strong performance, surpassing seven baseline approaches, highlighting its effectiveness for challenging SBDD tasks.
\end{itemize}

\section{Methodology}
\begin{figure*}[t]
\centering
\includegraphics[width=1\textwidth, height=9cm]{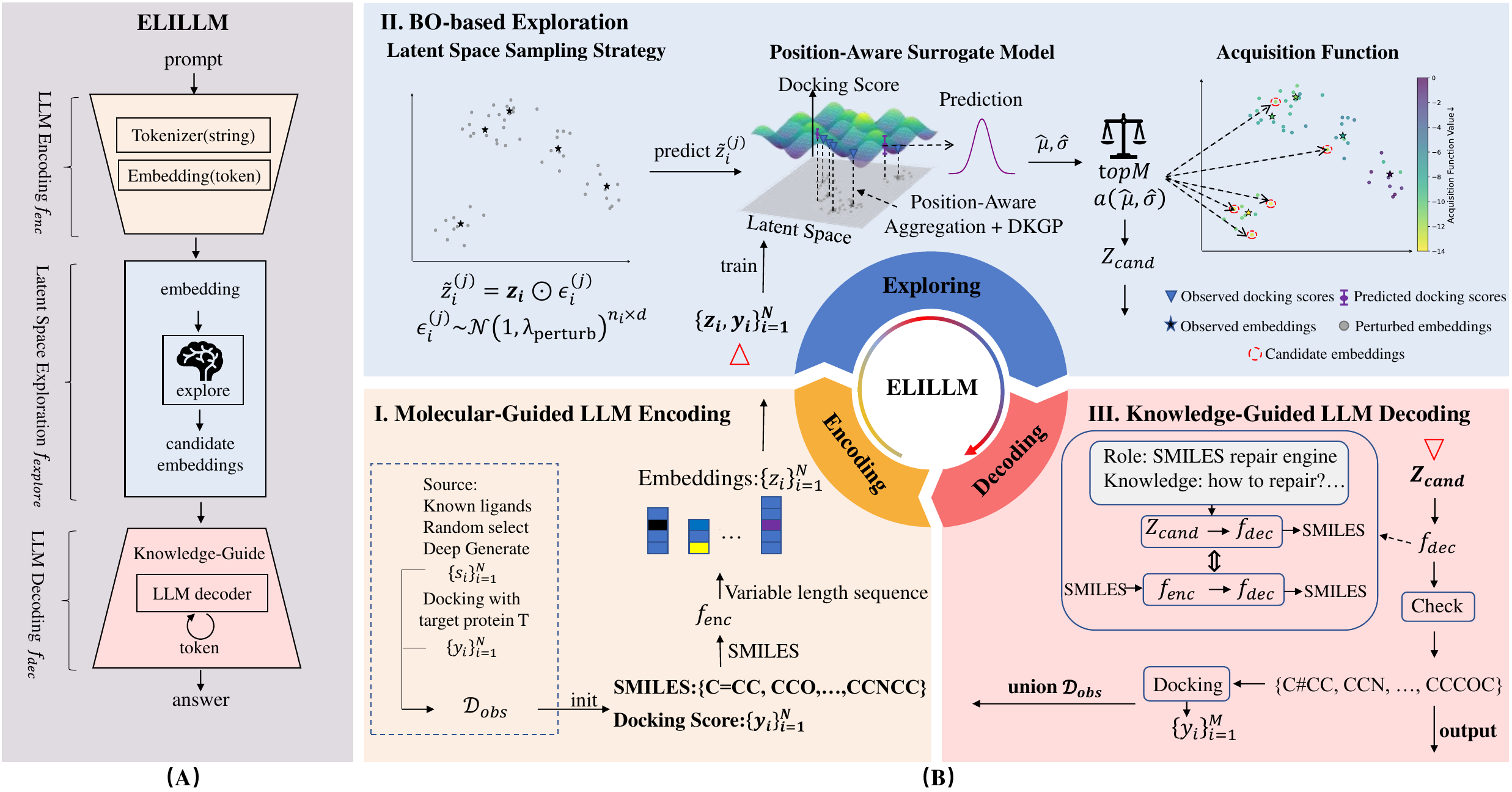}
\caption{Overview of the ELILLM framework. 
(A) The abstract framework of ELILLM. 
(B) A concrete instantiation of ELILLM, illustrating how the framework operates in a single iteration. Specifically, molecules in the observed dataset $D_\text{obs}$ are first encoded into embeddings. Next, the \textbf{Latent Space Sampling Strategy} samples a large set of candidate points to be evaluated. Simultaneously, the \textbf{Position-Aware Surrogate Model} is trained using the molecular embeddings as inputs and the corresponding docking scores as labels. The trained surrogate model is then used to predict the docking score distribution of the candidate points, and the acquisition function balances exploitation and exploration to select the most promising embeddings. Finally, through \textbf{Knowledge-Guided LLM Decoding}, chemical knowledge is used to constrain the LLM’s decoding behavior, converting candidate embeddings into the closest valid molecular SMILES to form a candidate molecule set. The candidate molecules are evaluated with black-box docking software, and the resulting molecule–score pairs are incorporated back into $D_\text{obs}$.}

\label{fig:framework}
\end{figure*}

In this section, we describe the framework of ELILLM, which enables exploration in the latent space of LLM. The structure of this framework is illustrated in Figure~\ref{fig:framework} (A).
To address SBDD tasks, we formulate the generation process of LLM as three main processes, including encoding, exploring, and decoding. Firstly, the encoding process is formulated as \( f_{\text{enc}} \): 
\begin{equation}
f_{\text{enc}}(s) = E[\text{tokenizer}(s)], \quad f_{\text{enc}}: \text{String} \rightarrow \mathbb{R}^{l \times d}.
\end{equation}

The \( \text{tokenizer}(s) \in \mathbb{N}^l \) is the tokenized input sequence, and \( E \in \mathbb{R}^{V \times d} \) is the embedding matrix that maps tokens to \( d \)-dimensional vectors. 
Then, we introduce the latent-space exploring process \( f_{\text{explore}} \), which can systematically explore a continuum of possibile entities with desired properties. 
It is defined as:
\begin{equation}
f_{\text{explore}}(\cdot): \mathcal{X} \rightarrow \mathbb{R}^{l' \times d},
\label{eq:fexplore}
\end{equation}
where \( \mathcal{X} \) denotes an unconstrained input space that can include prompt embeddings, prior knowledge, task-specific conditions, or any other information source. The only requirement is that the output of \( f_{\text{explore}} \) lies in the LLM's latent space, i.e., it must be a valid embedding in \( \mathbb{R}^{l' \times d} \). Since this process transforms one latent embedding into another through exploration and effectively performs a step of reasoning in the latent space, we refer to it as \emph{exploration-augmented latent inference}. Such embeddings can then be directly consumed by the decoder \( f_{\text{dec}} \) for autoregressive generation.
Next, the decoding process is denoted by \( f_{\text{dec}} \) as follows: \begin{equation}
f_{\text{dec}}(z) = \text{LLM}(z), \quad f_{\text{dec}}: \mathbb{R}^{l \times d} \rightarrow \text{String}.
\label{eq:fdec}
\end{equation}

The decoding process represents the autoregressive generation process of an LLM.  \( f_{\text{dec}} \)  consumes the embedding vectors \( z \in \mathbb{R}^{l \times d} \), and proceeds with autoregressive generation one token at a time, until an end-of-sequence token is produced. 
Overall, a single iteration of the ELILLM framework can be formalized as:
\begin{equation}
\text{ELILLM}(\text{prompt}) = f_{\text{dec}}\left(f_{\text{explore}}\left(f_{\text{enc}}(\text{prompt}),\; \star \right)\right),
\end{equation}
where \( \star \) denotes any additional input that conditions the exploration process, such as task-specific prior knowledge, optimization signals, or structural constraints. The decoder \( f_{\text{dec}} \) directly consumes the latent representation produced by \( f_{\text{explore}} \), and generates an output sequence autoregressively.

\section{Method Details }
In this section, we present the concrete implementation of the ELILLM framework. 
The detailed modeling based on the ELILLM framework are provided in the following. 
\subsection{Problem Definition}
SBDD plays a pivotal role in drug discovery, aiming to create ligand molecules with high affinity for a specific protein binding pocket.
In this work, we use \( s \in \mathbb{S} \) to denote a ligand molecule and \( t \in \mathbb{T} \) to represent the target protein and its binding pocket. To evaluate the binding affinity between \( s \) and \( t \), we employ a black-box docking score function \( \text{Dock}: \mathbb{S} \times \mathbb{T} \rightarrow \mathbb{R} \), which leverages the 3D structural information. A lower docking score indicates a stronger binding affinity. Given an observed dataset denoted as:
\begin{equation}
\mathcal{D}_{\mathrm{obs}} = \{(s_i, y_i)\}_{i=1}^N,
\end{equation}
where each \( s_i \in \mathbb{S} \) is a symbolic molecular sequence and \( y_i = \text{Dock}(s_i, t) \) is its associated docking score, our objective is to find a new subset \( \mathcal{S}_{\mathrm{top}} \subseteq \mathbb{S} \setminus \{s_i\}_{i=1}^N \) of size \( k \):
\begin{equation}
\mathcal{S}_{\mathrm{top}} = \underset{\mathcal{S}' \subseteq \mathbb{S}_{\mathrm{rem}}}{\operatorname{argmin}} \sum_{s \in \mathcal{S}'} \text{Dock}(s, t),
\end{equation}
where \( \mathbb{S}_{\mathrm{rem}} = \mathbb{S} \setminus \{s_i\}_{i=1}^N \) and \( |\mathcal{S}'| = k \).
\subsection{Overview}
In the SBDD task, we model three processes of the ELILLM framework, which is depicted in Figure~\ref{fig:framework}(B).
The process of Molecular-Guided LLM Encoding encodes observed molecules into the semantic latent space of the LLM, resulting in their corresponding latent embeddings. 
The process of BO-based Exploration in LLM Latent Embedding Space aims to identify embeddings that are expected to yield improved docking scores, which leverages the advantage of Bayesian optimization in efficiently searching black-box objectives under limited samples. It consists of three submodels, including Latent Space Sampling Strategy Model, Position-Aware Surrogate Model and Acquisition Function Model. Specifically, the Latent Space Sampling Strategy Model perturbs known molecular embeddings with multiplicative Gaussian noise to progressively generate a large set of exploratory points in each round based on the idea of spatial locality. The Position-Aware Surrogate Model employs a position-aware aggregation mechanism that is capable of handling variable-length embedding sequences while perceiving positional information, and efficiently computes the predictive distribution of binding affinities. Finally, the Acquisition Function Model exploits the predictive distribution to balance exploitation and exploration, selecting candidate embeddings with high exploration potential.
The process of Knowledge-Guided LLM Decoding guides the LLM to perform robust decoding by incorporating basic chemical knowledge and defining its role as a SMILES repair engine. Each candidate embedding is decoded into a valid molecular SMILES, evaluated using the black-box docking score, and added to the observed data to complete one iteration. The pseudocode is provided in Appendix~A.


\subsection{Molecular-Guided LLM Encoding}
We first introduce the construction of observed dataset. For targets with known high-affinity ligands, existing ligands can be directly utilized.  In novel drug discovery scenarios, we may either randomly sample molecules from existing chemical libraries or employ existing deep generative models as tools to produce initial ligand candidates that satisfy basic requirements for the given target protein. 

Given the observed dataset \(\mathcal{D}_{\text{obs}} = \{(s_i, y_i)\}_{i=1}^N\), we compute latent embeddings \(z_i = f_{\text{enc}}(s_i) \in \mathbb{R}^{n_i \times d},\)
where \(n_i\) denotes the length of the token sequence corresponding to \(s_i\), and \(d\) is the hidden dimension of the embedding space. 
The set of latent embeddings of observed molecules is denoted as:
\begin{equation}
\label{eq:zobs}
\mathcal{Z}_{\text{obs}} = \{ z_i \}_{i=1}^{N}.
\end{equation}
\subsection{BO-based Exploration}
\subsubsection{Latent Space Sampling Strategy.}

To enable local exploration in the latent space, we apply multiplicative Gaussian perturbations to each \(z_i\), generating perturbed embeddings via element-wise noise:
\begin{equation}
\label{eq:perturb}
\tilde{z}_i^{(j)} = z_i \odot \epsilon_i^{(j)}, 
\epsilon_i^{(j)} \sim \mathcal{N}(1, \lambda_{\text{perturb}})^{n_i \times d},
\end{equation}
where \(j = 1, \dots, M\). \(\lambda_{\text{perturb}}\) is a tunable hyperparameter that controls the magnitude of perturbation. A larger \(\lambda_{\text{perturb}}\) results in wider exploration around the original embedding, while a smaller value enforces more localized sampling. The set of perturbed embeddings is denoted by:
\begin{equation}
\label{eq:zexplore}
\mathcal{Z}_{\text{explore}} = \left\{ \tilde{z}_i^{(j)} \;\middle|\; i = 1, \dots, N; \; j = 1, \dots, M \right\}.
\end{equation}


\subsubsection{Position-Aware Surrogate Model.}
While recent studies have proposed expressive molecular representation learning methods based on graph neural networks~\cite{rf:csgl,rf:reaction_graph}, these approaches are primarily designed for supervised or self-supervised learning on explicit molecular graphs.
In contrast, our surrogate model operates directly on the latent representations of a pretrained LLM and must be trained online with a limited number of evaluated samples, where reliable uncertainty estimation and sample efficiency are critical.
As a result, we adopt a lightweight, position-aware surrogate that aligns with the requirements of iterative black-box optimization in latent space.

A key requirement for the surrogate model is to provide well-calibrated uncertainty estimates, which are essential for balancing exploration and exploitation during latent space optimization.
Gaussian Processes (GPs)~\cite{rf:gp} are a natural choice due to their principled Bayesian formulation and inherent uncertainty quantification.
However, modern LLMs such as LLaMA~3.1~\cite{llama3.1} typically produce high-dimensional molecular embeddings (e.g., \( d = 4096 \)), which pose significant challenges for standard GPs.
In such high-dimensional spaces, data sparsity severely degrades kernel expressiveness, leading to unreliable predictive uncertainty~\cite{rf:gplimit}.

To address these challenges, we employ a Deep Kernel Gaussian Process (DKGP)~\cite{rf:dkgp} as our surrogate model.
DKGP integrates a learnable neural projection that maps high-dimensional LLM embeddings into a lower-dimensional, task-adaptive feature space, enabling effective uncertainty-aware modeling while preserving the Bayesian advantages of GPs.
This design allows the surrogate to remain both expressive and sample-efficient under the constraints of online latent-space optimization.

Additionally, each molecular embedding $\widetilde{z}_i^{(j)} \in \mathbb{R}^{n_i \times d}$ is a variable-length sequence, where $n_i$ differs across molecules. 
Although models like Transformers and LSTMs are capable of directly processing such sequences, their computational demands become impractical for the evaluation of thousands of candidates in iterative searches. 

To preserve positional information during aggregation, we compute a position-aware embedding as:
\begin{equation}
\bar{z}_i^{(j)} = \frac{1}{n_i} \sum_{t=1}^{n_i} \frac{
\mathrm{concat}\big(p_t \widetilde{z}_{i,t}^{(j)},\ (l_{\max} - p_t) \widetilde{z}_{i,t}^{(j)}\big)
}{l_{\max}},
\label{eq:aggregation}
\end{equation}
where $p_t$ denotes the positional index of the token embedding $\widetilde{z}_{i,t}^{(j)}$, and $l_{\max}$ is the predefined maximum number of generated tokens per molecule. Thus, $\bar{z}_i^{(j)}$  encodes both the absolute and relative position of each token, which can enhance the surrogate model’s ability to capture structural information in the latent space. 

Our position-aware aggregation introduces a novel form of position sensitivity by reweighting token embeddings according to their positions. Moreover, under our setting where token embeddings are obtained without positional encoding (Theorem~\ref{thm:embedding_indep}), 
the embeddings are independent of their positions. Combined with the proposed aggregation operation, this ensures that the resulting aggregated embeddings are positionally unbiased (Theorem~\ref{thm:unbiased}). The detailed proofs are provided in Appendix~\ref{app:proof}.


\begin{theorem}[Independence of Token Embeddings and Positions]
\label{thm:embedding_indep}
Let $s = (x_1, x_2, \dots, x_l)$ be a token sequence, and let 
$\mathrm{tokenizer}(s)$ map $s$ to discrete token indices $(t_1, t_2, \dots, t_l)$. 
Let $E$ be an embedding matrix mapping each token $t$ to a vector $E[t] \in \mathbb{R}^d$, and define the token embeddings as
\[
\widetilde{z}_{t} = E[t].
\]
If no positional encoding is added (e.g., no sinusoidal or learned positional embedding), 
then for each sequence $s$ the embedding $\widetilde{z}_{t}$ of token $t$ is independent of its positional index $p_t \in \{1, \dots, l\}$.
\end{theorem}

\begin{theorem}[Positional Unbiasedness of Aggregation]
\label{thm:unbiased}
Assume the token embeddings $\widetilde{z}_{i,t}^{(j)}$ are obtained via a tokenizer and embedding matrix 
without any positional encoding, so that by Theorem~\ref{thm:embedding_indep}, 
they are independent of their positional indices $p_t$. 
Then, the position-aware aggregation defined in Eq.~\eqref{eq:aggregation} 
is statistically unbiased with respect to position, i.e., 
the expected aggregated embedding $\mathbb{E}[\bar{z}_i^{(j)}]$ treats all token positions equally.
\end{theorem}


The aggregated embedding $\bar{z}_i^{(j)} \in \mathbb{R}^{2d}$ is then passed through a multilayer perceptron (MLP), yielding a transformed feature vector:
\begin{equation}
\hat{z}_i^{(j)} = f_{\mathrm{MLP}}(\bar{z}_i^{(j)}).
\end{equation}

We denote this overall transformation as \( \phi_{\boldsymbol{\theta}}(\cdot) \), including the position-aware aggregation and dimensionality-reducing through an MLP. 
\( \phi_{\boldsymbol{\theta}}(\cdot) \) maps a variable-length embedding sequence to a fixed-length vector: \( \phi_{\boldsymbol{\theta}}: \mathbb{R}^{n_i \times d} \rightarrow \mathbb{R}^{d'} \), where \( d \) is the embedding dimension (e.g., 4096) and \( d' \) is a predefined lower dimension (e.g., 20).
The GP is defined over the transformed inputs, giving the surrogate model:
\begin{equation}
\label{eq:predict}
\hat{f_{\boldsymbol{\theta}}}(z) \sim \mathcal{GP}(0, k_{\mathrm{GP}}(\phi_{\boldsymbol{\theta}}(z), \phi_{\boldsymbol{\theta}}(z'))),
\end{equation}
where $k_{\mathrm{GP}}$ is the learnable kernel function in the transformed space and the detailed construction of the kernel function is provided in Appendix~\ref{app:method}.
For any explored embeddings $\tilde{z}_i^{(j)}$, the model produces a predictive distribution with mean $\hat{\mu}(\tilde{z}_i^{(j)})$ and variance $\hat{\sigma}^2(\tilde{z}_i^{(j)})$, representing the predicted docking score and the model uncertainty, respectively:
\begin{equation}
\hat{f}_{\boldsymbol{\theta}}\left(\tilde{z}_i^{(j)}\right) \sim \mathcal{N}\left(\hat{\mu}\left(\tilde{z}_i^{(j)}\right),\; \hat{\sigma}^2\left(\tilde{z}_i^{(j)}\right)\right)
\end{equation}
\subsubsection{Hierarchical Training Strategy.}
We adopt a hierarchical training strategy for the surrogate model. The training dataset \( \mathcal{D}_{\text{train}} = \{(z_i, y_i)\}_{i=1}^{N} \) is constructed from the latent embeddings \( \mathcal{Z}_{\text{obs}} \) and their associated docking scores.

In the first stage, we train a composite predictor that maps each latent input \( z_i \) to a scalar prediction using a transformation \( \phi_{\boldsymbol{\theta}} \) followed by a regression head \( h \). The entire model is optimized by minimizing the mean squared error loss:
\begin{equation}
\label{eq:mse}
\mathcal{L}_{\mathrm{MSE}} = \frac{1}{N} \sum_{i=1}^N \left( h\left(\phi_{\boldsymbol{\theta}}(z_i)\right) - y_i \right)^2.
\end{equation}

After convergence, the regression head is discarded and \( \phi_{\boldsymbol{\theta}} \) is frozen. In the second stage, a Gaussian process is trained on the transformed embeddings by minimizing the negative log marginal likelihood:
\begin{equation}
\label{eq:nll}
\mathcal{L}_{\mathrm{GP}} = - \log p(\mathbf{y} \mid \phi_{\boldsymbol{\theta}}(\mathbf{Z})),
\end{equation}
where \( \mathbf{Z} = \mathcal{Z}_{\text{obs}} \) and \( \mathbf{y} = \{y_i\}_{i=1}^N \).

This two-stage design decouples representation learning from uncertainty modeling, allowing the model to first focus on effective feature extraction and then on principled uncertainty quantification. Such separation enhances training stability and improves predictive performance, particularly under the data-scarce conditions commonly encountered in surrogate modeling.

\subsubsection{Acquisition Function.}
We adopt the lower confidence bound (LCB) acquisition function to guide exploration in the latent space. This acquisition function balances the predictive mean and uncertainty of the surrogate model, and is defined as:
\begin{equation}
\alpha_{\mathrm{LCB}}(z) = \hat{\mu}(z) - \kappa_t \cdot \hat{\sigma}(z),
\end{equation}
where \( \hat{\mu}(z) \) and \( \hat{\sigma}(z) \) denote the predictive mean and standard deviation, respectively, predicted by the surrogate model given the latent input \( z \).

We follow the schedule proposed by Srinivas et al.~\cite{srinivas2010gaussian}, due to its strong theoretical guarantees in the Bayesian optimization setting. Specifically, the exploration coefficient \( \kappa_t \) is dynamically updated as:
\begin{equation}
\kappa_t = \sqrt{2 \log \left( \frac{t^2 \pi^2}{6 \delta} \right)},
\end{equation}
where \( \delta \in (0, 1) \) is a confidence parameter. In our setup, we set \( t = |\mathcal{D}_{\text{train}}| + 1 \), such that the exploration strength naturally increases with the amount of observed data.


After computing the acquisition scores for all perturbed latent embeddings \( \mathcal{Z}_{\mathrm{explore}} \), we select the top candidates for further evaluation. Specifically, we choose the \( n_{\mathrm{cand}} \) latent embeddings with the lowest acquisition scores, forming the candidate set:
\begin{equation}
\label{eq:zcand}
\mathcal{Z}_{\mathrm{cand}} = \underset{z \in \mathcal{Z}_{\mathrm{explore}}}{\operatorname{argmin\_top}^{n_{\mathrm{cand}}}} \, \alpha_{\mathrm{LCB}}(z).
\end{equation}

\renewcommand{\arraystretch}{0.88}
\begin{table*}[t]
\centering
\caption{Evaluation of generated molecules is primarily based on the \textbf{Vina docking score} (Top 1, 5, 10, and 20), with \textit{QED}, \textit{SA}, and \textit{diversity} serving as auxiliary metrics. We report both average (Avg.) and median (Med.) values across targets; for most metrics, Med.\ denotes the mean of per-target medians, while for diversity—yielding one value per target—Med.\ denotes the median of the 100 target-level scores. The best and second-best results are highlighted in \textbf{bold} and \underline{underlined}, respectively.}
\setlength{\tabcolsep}{1.4mm}
\begin{tabular}{l
  cc|cc|cc|cc|cc|cc|cc}
\toprule
\textbf{Methods} 
& \multicolumn{2}{c|}{\textbf{Top1 Dock} $\downarrow$} 
& \multicolumn{2}{c|}{\textbf{Top5 Dock}$\downarrow$} 
& \multicolumn{2}{c|}{\textbf{Top10 Dock}$\downarrow$} 
& \multicolumn{2}{c|}{\textbf{Top20 Dock}$\downarrow$} 
& \multicolumn{2}{c|}{\textit{QED}} 
& \multicolumn{2}{c|}{\textit{SA}} 
& \multicolumn{2}{c}{\textit{Diversity}} \\
& Avg. & Med. & Avg. & Med. & Avg. & Med. & Avg. & Med. & Avg. & Med. & Avg. & Med. & Avg. & Med. \\
\midrule
Random Init & -8.62 & -8.62 & -8.19 & -8.14 & -7.90 & -7.81 & -6.38 & -7.42 & 0.53 & 0.53 & 0.76 & 0.77 & 0.78 & 0.77 \\
AR (2021)   & -8.41 & -8.41 & -8.08 & -8.03 & -7.88 & -7.82 & -7.64 & -7.56 & 0.51 & 0.50 & 0.63 & 0.63 & 0.70 & 0.70 \\
Pocket2Mol (2022) & -9.15 & -9.15 & -8.85 & -8.82 & -8.64 & -8.60 & -8.38 & -8.32 & 0.56 & 0.57 & 0.74 & 0.75 & 0.69 & 0.71 \\
liGAN (2022)     & -8.13 & -8.13 & -7.78 & -7.73 & -7.58 & -7.53 & -7.34 & -7.26 & 0.39 & 0.39 & 0.59 & 0.57 & 0.66 & 0.67 \\
TargetDiff (2023)& \underline{-9.38} & \underline{-9.38} & -8.86 & -8.78 & -8.53 & -8.43 & -8.13 & -8.00 & 0.48 & 0.48 & 0.58 & 0.58 & 0.72 & 0.71 \\
ALIDIFF (2024)  & -9.37 & -9.37 & -8.85 & -8.80 & -8.52 & -8.40 & -8.11 & -8.01 & 0.50 & 0.50 & 0.57 & 0.56 & 0.73 & 0.71 \\
TamGen (2024)     & -8.53 & -8.53 & -8.11 & -8.05 & -7.72 & -7.65 & -7.70 & -7.62 & 0.56 & 0.56 & 0.77 & 0.78 & 0.75 & 0.74 \\
LMLF-rand (2024) & -8.66 & -8.66 & -8.17 & -8.10 & -7.87 & -7.78 & -7.50 & -7.40 & 0.57 & 0.59 & 0.77 & 0.79 & 0.81 & 0.80 \\
LMLF-diff (2024)   & -9.05 & -9.05 & -8.52 & -8.43 & -8.19 & -8.11 & -7.79 & -7.67 & 0.53 & 0.53 & 0.64 & 0.63 & 0.78 & 0.77 \\ \hline
\textbf{ELILLM-rand} & -9.33 & -9.33 & \underline{-8.98} & \underline{-8.93} & \underline{-8.74} & \underline{-8.68} & \underline{-8.44} & \underline{-8.36} & 0.46 & 0.45 & 0.63 & 0.63 & 0.69 & 0.70 \\
\textbf{ELILLM-diff}   & \textbf{-9.80} & \textbf{-9.80} & \textbf{-9.37} & \textbf{-9.31} & \textbf{-9.09} & \textbf{-9.03} & \textbf{-8.74} & \textbf{-8.65} & 0.49 & 0.49 & 0.57 & 0.56 & 0.67 & 0.68 \\
\bottomrule
\end{tabular}
\label{tab:mainresult}
\end{table*}

\subsection{Knowledge-Guided LLM Decoding}
The current problem is how to design the LLM decoding function \( f_{\text{dec}} \) that transforms \( Z_{\text{cand}} \) into valid ligand molecules.
However, reliably implementing this function in practice presents significant challenges. Due to the randomness of LLM generation, the same input can yield different outputs if not properly constrained, leading to instability in decoding. Moreover, since candidate embeddings are derived via E.q.~\ref{eq:perturb}, they may partially lie outside the valid chemical embedding space \( \mathcal{Z}_s \), increasing the likelihood of producing invalid or off-target molecules.

\begin{figure*}[t]
\centering
\includegraphics[width=1\textwidth,height=3.2cm, keepaspectratio=false]{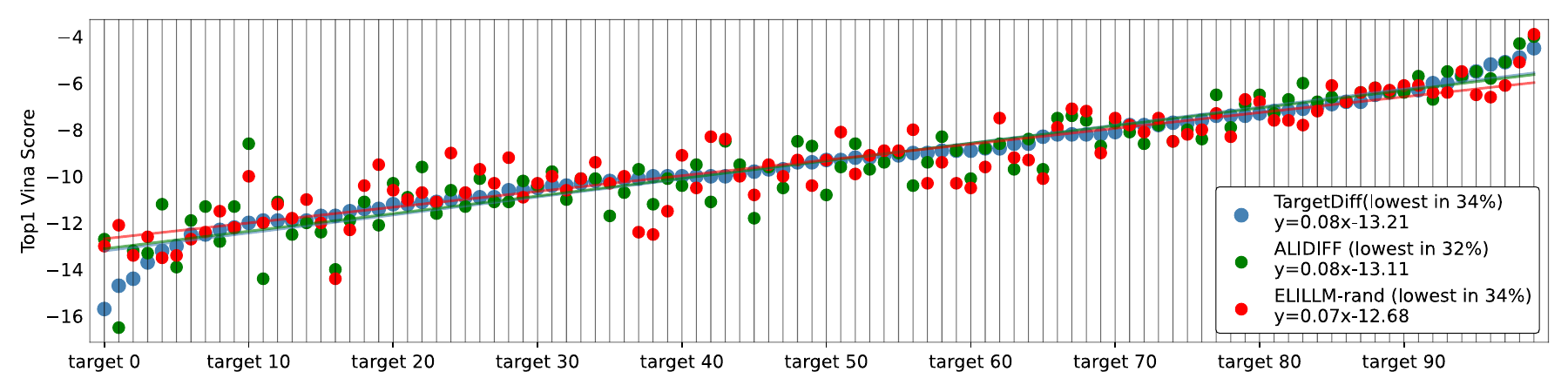}
\caption{Top1 Vina docking score for different generated molecules (TargetDiff, ALIDIFF, ELILLM-rand) across
100 testing targets, sorted by Vina docking score of TargetDiff result. Trend lines are least-squares linear fits for each method.}
\label{fig:scatter_per_target}
\end{figure*}
To address these issues, we employ knowledge-guided, role-based prompting to guide the LLM. First, we define the LLM's role as a SMILES repair engine, prompting it to minimally adjust candidate embeddings into chemically plausible structures. Building on this, we provide chemically informed instructions (e.g. complete missing ring numbers: C1CC → C1CC1) to handle common error types, further constraining the generation process. The entire decoding process is formulated as follows:

\begin{equation}
\label{eq:scand}
\mathcal{S}_{\mathrm{cand}} = \{ f_{\mathrm{dec}}(z,\text{prompt}) \mid z \in \mathcal{Z}_{\mathrm{cand}} \},
\end{equation}
where \( \mathcal{S}_{\mathrm{cand}} \) denotes the set of SMILES strings generated from the latent candidates. Afterwards, we will store \( \mathcal{S}_{\mathrm{cand}} \) and their docking scores in \( \mathcal{D}_{\mathrm{obs}}\) to guide the generation of subsequent iterations.
\begin{equation}
\mathcal{D}_{\mathrm{obs}} \leftarrow \mathcal{D}_{\mathrm{obs}} \cup \{(s, \text{Dock}(s, t)) \mid s \in \mathcal{S}_{\mathrm{cand}}\}
\end{equation}

\section{Experiments}
In this section, we evaluate our ELILLM on the SBDD task to answer two main questions: 
(Q1) can the ELILLM framework achieve the expected generation performance on domain-specific tasks? 
(Q2) how does ELILLM perform efficient exploration in high-affinity regions?
In addition, we conduct ablation studies and visualization experiments to further demonstrate the effectiveness of our framework. Owing to space constraints, supplementary experiments (e.g., the Wilcoxon signed-rank test) are included in Appendix~\ref{app:addexp}.

\begin{figure*}[!t]
\centering
\includegraphics[width=1\textwidth,height=3.2cm, keepaspectratio=false]{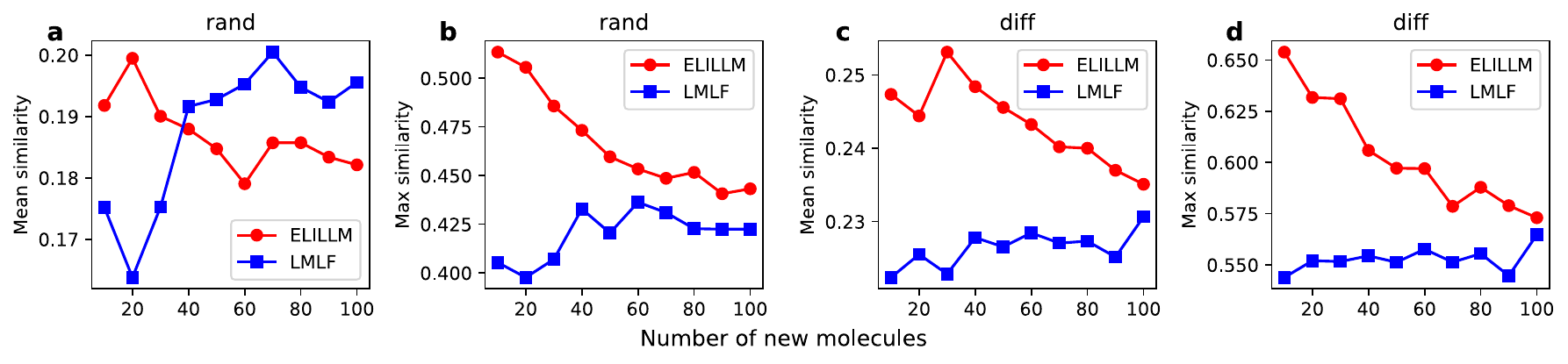}
\caption{Average Tanimoto similarity between every 10 newly generated molecules and the initial \(\mathcal{D}_{\mathrm{obs}}\), including both the average of all pairwise similarities (\textit{mean similarity}) and the average of per-molecule maximum similarity to \(\mathcal{D}_{\mathrm{obs}}\) (\textit{max similarity}). Results are averaged across 100 targets to show overall trends under two \(\mathcal{D}_{\mathrm{obs}}\) construction strategies.}
\label{fig:similarity_curve}
\end{figure*}
\subsection{Experiment Setup}
\subsubsection{Dataset and Settings.}
We evaluate ELILLM using the CrossDocked2020 dataset\cite{rf:crossdocked2020}. In line with standard practice~\cite{rf:alidiff, ref:tamgen}, the dataset includes 65K drug--target pairs for training and 100 target proteins for testing. Unlike most methods that train on the entire training set, we consider two settings for constructing the observed dataset \( \mathcal{D}_{\text{obs}} \): (1) randomly selecting 100 ligands from the training set, docking them against the test targets, and using the resulting ligand--score pairs; and (2) directly adopting the sampled ligands and their docking scores against the 100 test targets as provided by ALIDIFF~\cite{rf:alidiff}.  
The first setting, referred to as \textbf{ELILLM-rand}, is designed to evaluate the effectiveness of our method when initialized from a random chemical space. The second setting, termed \textbf{ELILLM-diff}, demonstrates that our framework can be effectively integrated with existing deep learning–based generation approaches. Meanwhile, based on the conclusions of \citet{rf:usellama}, we choose to use LLaMAs 3.1 as the basic LLM underlying both our method and related baseline approaches.

\subsubsection{Baselines.}
We compare ELILLM with the following baselines: 
\textbf{Random Init} refers to the randomly constructed \( \mathcal{D}_{\text{obs}} \) mentioned earlier, which is used by both ELILLM-rand and LMLF-rand. \textbf{liGAN}~\cite{rf:ligan} is a conditional VAE model based on 3D voxelized representations. \textbf{AR}~\cite{rf:ar} and \textbf{Pocket2Mol}~\cite{rf:pocket2mol} are 3D autoregressive models at the atom level. \textbf{TargetDiff}~\cite{rf:targetdiff} and \textbf{ALIDiff}~\cite{rf:alidiff} are diffusion-based approaches designed for target-conditioned molecular generation. Among them, ALIDiff is a recent SOTA method that fine-tunes a pretrained diffusion model through preference optimization.\textbf{TamGen}~\cite{ref:tamgen} is a GPT-based chemical language model that generates ligand SMILES conditioned on a given target protein. \textbf{LMLF}~\cite{rf:lmlf} is a method that leverages logical feedback to constrain a general-purpose LLM for SBDD, producing ligand molecules in the form of SMILES string under the given \( \mathcal{D}_{\text{obs}} \). Correspondingly, we also implement a variant of LMLF, referred to as \textbf{LMLF-diff}.

\subsubsection{Evaluation Metrics.}
To evaluate the quality of generated molecules in the context of virtual screening, we generate 100 candidate molecules for each of the 100 target proteins in the test set. The generated molecules are evaluated primarily in terms of binding affinity, estimated by the Vina docking score, which directly reflects the LLM’s targeted generation ability. Importantly, in our experiments, only the Vina docking score is used as the optimization guidance, while auxiliary metrics, including drug-likeness (QED) \cite{rf:qed}, reversed normalized synthetic accessibility (SA) \cite{rf:sa}, and molecular diversity, are reported for reference to illustrate general molecular properties. Internal diversity, which can be simply measured as the average pairwise Tanimoto distance between generated molecules, may slightly decrease when exploration is concentrated in regions of high docking affinity. In our experiments, the diversity remains above 0.5, which is considered a reasonable range to ensure meaningful coverage of chemical space. QED and SA often exhibit a negative correlation with docking scores and therefore cannot serve as direct indicators of targeted binding affinity.

To ensure a fair comparison, we eliminate the influence of varying 3D coordinates. 3D deep generative models may produce different conformations for the same molecule, potentially leading to inconsistent docking scores. Therefore, we extract the SMILES representation of each generated molecule and reconstruct its 3D conformation using RDKit’s rule-based coordinate generation. The conformations are then optimized via a molecular force field to improve geometric quality. The resulting structures are used for docking with smina to estimate binding affinity scores. In this setup, by controlling the random seed, we ensure that the same SMILES corresponds to the same Vina Docking Score.

Due to space limitations, more detailed experimental settings are provided in Appendix~\ref{app:expsetting}.
\subsection{Performance of SBDD}
To answer Q1, we compare ELILLM with baseline methods in a simplified virtual screening scenario, aiming to identify candidate sets of size 1, 5, 10, and 20 with the lowest Vina docking scores for subsequent lead optimization. 
The results are presented in Table~\ref{tab:mainresult}. As shown, \textbf{ELILLM-rand} outperforms all baseline methods in SBDD, achieving superior performance under the Top~5, 10, and 20 candidate settings despite using only 100 randomly selected \(\mathcal{D}_{\mathrm{obs}}\) samples from the full 65K training dataset. While ELILLM-rand demonstrates strong performance, ELILLM-diff further achieves SOTA binding affinity by integrating our framework with a pretrained diffusion model. Specifically, ALIDiff serves as the base generative model. ELILLM-diff outperforms ALIDiff by 4.59\%, 5.88\%, 6.69\%, and 7.77\% in the Top 1, 5, 10, and 20 candidate settings, respectively, in terms of binding affinity, showcasing the strong compatibility and enhancement our framework brings to existing deep generative approaches. Moreover, as shown in Appendix Table~\ref{tab:docking_pvalues}, a Wilcoxon signed-rank test significantly rejects the one-sided hypotheses that ELILLM-diff is weaker than TargetDiff and ALIDIFF (p < 0.01), indicating that our results are statistically significant.

Notably, although LMLF achieves lower binding affinity compared to the provided molecules, it shows consistently strong performance in terms of QED, SA, and molecular diversity. This pattern reflects a commonly observed empirical trade-off: QED and SA scores often exhibit a negative correlation with binding affinity. From an encoder–decoder perspective, QED, SA, and diversity fall within the part of the task that LLMs can handle natively via their internal representations and decoding strategies. In contrast, optimizing docking affinity corresponds to the component that lies beyond the LLM’s direct capability and is addressed by our ELILLM framework, which performs controlled exploration in the latent space guided by domain knowledge. The entities generated through this exploration effectively act as intermediate reasoning steps that augment the LLM’s decoding, enabling more targeted generation of high-affinity molecules.
These results underscore the importance of incorporating domain knowledge to guide the reasoning and exploration process in the LLM latent space. 

As shown in Figure~\ref{fig:scatter_per_target}, although the mean Top 1 Vina docking score of ELILLM-rand is slightly higher than those of ALIDIFF and TargetDiff, the three methods actually demonstrate comparable performance under this setting. 
We further observe that the two diffusion-based methods produce nearly identical trend lines, likely due to reliance on the same training distribution. By contrast, our methods, conditioning on a small number of molecules, enable more flexible generation with potential advantages on targets possibly dissimilar to the training set (e.g., Targets 79 and 96).

\subsection{Performance of Chemical Space Exploration}
To answer Q2, we aim to evaluate whether our method progressively explores the chemical space or merely remains confined to the region anchored by the initial 
\(\mathcal{D}_{\mathrm{obs}}\).
We measure the distance between generated molecules and the initial space using changes in Tanimoto similarity\cite{rf:tanimoto} during the iterative generation process.
Figure~\ref{fig:similarity_curve} illustrates the chemical space exploration capabilities of our method and LMLF in two settings. 
It is observed that our method generates molecules progressively farther from the initial space, whereas LMLF is significantly limited to the initial region and even tends to move closer over iterations, despite achieving the highest diversity. Furthermore, by jointly examining subfigures (a) and (b), we observe that in the early stages of generation, our method exhibits an increase in mean similarity while the max similarity decreases, followed by a consistent downward trend in both metrics. 
These observations indicate that ELILLM behaves like a human expert: initially leveraging existing knowledge to design new molecules by moving away from the most similar instances and instead exploring directions aligned with other diverse examples. It then dynamically incorporates newly acquired knowledge to discover new directions for molecular design.
\renewcommand{\arraystretch}{0.8}
\begin{table}[t]
\centering
\caption{Ablation study on the first 10 targets from the test set using ALIDIFF-generated \(\mathcal{D}_{\mathrm{obs}}\)}
\setlength{\tabcolsep}{1.4mm}
\begin{tabular}{l
  cc|cc|cc}
\toprule
\textbf{Methods} 
& \multicolumn{2}{c|}{\textbf{Top1 Dock} $\downarrow$} 
& \multicolumn{2}{c|}{\textbf{Top20 Dock}$\downarrow$} 
& \multicolumn{2}{c}{\textit{Diversity}} \\
& Avg. & Med. & Avg. & Med. & Avg. & Med. \\
\midrule
ELILLM   & \textbf{-9.86} & \textbf{-9.86} & \textbf{-8.79} & \textbf{-8.71} & 0.69 & 0.69\\
w/o guide & -9.08 & -9.08 & -8.01 & -7.90 & \textbf{0.76} & \textbf{0.75} \\
w/o position   & -9.80 & -9.80 & -8.52 & -8.38 & 0.73 & 0.72 \\
w/o knowledge   & -9.40 & -9.40 & -8.30 & -8.04 & 0.74 & 0.74 \\
w/o role & -8.20 & -8.20 & - & - & 0.48 & 0.60\\
ALIDIFF & -9.10 & -9.10 & -8.17 & -8.04 & 0.75 & 0.72 \\

\bottomrule
\end{tabular}
\label{tab:ablation}
\end{table}
\subsection{Ablation Study}
To better understand the contribution of each component in our framework, we conduct an ablation study on the first 10 test targets using ALIDiff-generated \(\mathcal{D}_{\mathrm{obs}}\) as initial observations. We evaluate four variants: w/o guide, w/o position, w/o role, and w/o knowledge. The w/o guide variant randomly samples \(\mathcal{Z}_{\mathrm{cand}}\) from \(\mathcal{Z}_{\text{explore}}\); w/o position uses simple average aggregation with concatenated vectors; w/o role has the LLM output the input string directly; and w/o knowledge includes role information only, without chemical knowledge.

As shown in Table~\ref{tab:ablation}, w/o guide fails to achieve targeted generation, highlighting the importance of our guidance strategy. w/o position performs slightly worse, confirming that position-aware aggregation captures latent-space patterns more effectively. w/o role exhibits unacceptable performance, generating fewer than 20 distinct molecules across all 10 targets, emphasizing the necessity of role-specific prompting for effective exploration. w/o knowledge shows weaker targeted generation but higher diversity, indicating that prior knowledge helps constrain LLM decoding towards high-affinity regions.


\subsection{Visualization}
\begin{figure}
    \centering
    \includegraphics[width=1\linewidth, height=3cm]{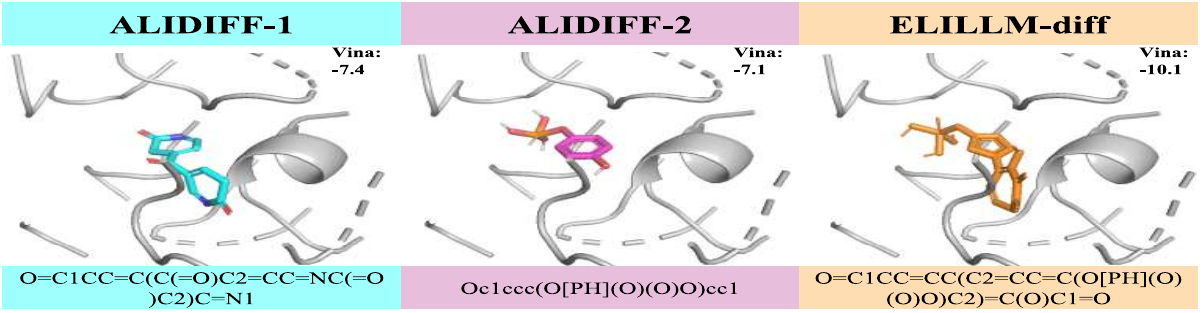}
    \caption{Visualizations of generated ligands for protein pockets 2jjg generated by ALIDIFF and ELILLM-diff.}
    \label{fig:visualization}
\end{figure}

We visualize the generated ligand for the protein pocket 2jjg to further demonstrate our expert-like exploratory generation process. As shown in Figure~\ref{fig:visualization}, without any human guidance (such as identifying key functional groups or scaffolds), our method explores the latent space and designs a new molecule that combines the “O=C1CC=C” fragment from ligand 1 and the “O[PH](O)(O)O” group from ligand 2, and completes the rest using other learned structural knowledge.




\section{Related Work}
\paragraph{LLM generation as encoder--decoder pipelines.}
LLM-based generation can be viewed as an encoder--decoder process, where the encoder maps inputs into latent representations and the decoder autoregressively generates outputs.
Prior work mainly enhances either the encoder, e.g., through knowledge-enhanced pretraining or soft prompts~\cite{rf:kepler,rf:dkplm,rf:softprompt1,rf:spot}, or the decoder via prompt engineering and feedback-based refinement~\cite{rf:cot,rf:sbs,rf:llmfunsearch,rf:lmlf}.
In contrast, our method enables explicit exploration and control within the latent space constructed by the encoder, supporting more structured and efficient scientific generation.

\paragraph{SBDD and LLM.} With the increasing availability of protein–ligand structural data, structure-based drug design (SBDD) has become a central application scenario for generative molecular modeling, with a wide range of 3D-aware generative approaches proposed in prior work. 
LLMs have shown promising capabilities across a variety of tasks beyond molecular generation, including reasoning, graph-based learning, and iterative feedback-guided optimization~\cite{rf:llmfunsearch,rf:noisy_llm_graph,rf:huangcode}.
These successes highlight the versatility of pretrained LLMs as knowledge sources that can be leveraged without task-specific finetuning.
 Building on this idea, models such as MolGPT~\cite{rf:molgpt}, LMLF~\cite{rf:lmlf}, Token-Mol~\cite{rf:tokenmol}, TamGen~\cite{ref:tamgen} and DrugLLM~\cite{rf:drugllm} leverage textual and structural representations learned during pretraining to generate molecules. However, these methods often require task-specific finetuning or large domain datasets to capture complex protein-ligand interactions effectively.
In contrast, our framework addresses this challenge in a \emph{plug-in} fashion, without retraining the LLM. ELILLM explicitly explores the latent space of pretrained LLMs using domain knowledge and a position-aware surrogate model to handle aspects of SBDD that the model alone cannot fully capture. By performing controlled latent space exploration, ELILLM generates candidate embeddings that augment the LLM’s decoding process, enabling targeted molecular generation while preserving the pretrained knowledge.

\section{Conclusion}
In this work, we present ELILLM, a novel exploration-augmented latent inference framework tailored for SBDD. ELILLM reformulates the LLM generation process into a three-stage pipeline of encoding, exploration, and decoding, explicitly leveraging the expressive power of latent spaces to address the challenge of systematic latent space navigation. We instantiate this framework for SBDD with three key components: Molecular-Guided LLM Encoding, BO-based Exploration, and Knowledge-Guided LLM Decoding.

Our evaluation on the CrossDocked2020 benchmark demonstrates that ELILLM achieves superior performance in binding affinity scores, surpassing seven baseline methods. Importantly, ELILLM shows strong compatibility with pretrained models, enabling it to be seamlessly integrated with existing LLM-based molecular generation approaches. By simulating human-like research workflows with interpretability and efficiency, ELILLM offers a principled and effective approach for optimizing docking-based molecular design.

Future work will focus on generalizing the proposed techniques to broader scenarios beyond SBDD. In particular, we aim to explore applications in graph representation learning~\cite{rf:dnea,rf:graphiam} and recommender systems~\cite{rf:diffrec,rf:hyperkg}, where latent space exploration and uncertainty-aware modeling could provide similar benefits.

\begin{acks}
This work is supported by the National Natural Science Foundation of China under Grant Nos. U22A2098, 62172185, 62206105, 62202200, and 62406127; the KeyScience and Technology Development Plan of Jilin Province under Grant No.20240302078GX.
\end{acks}

\bibliographystyle{ACM-Reference-Format}
\balance
\bibliography{sample-base}

\appendix

\renewcommand{\thefigure}{S\arabic{figure}}
\renewcommand{\thetable}{S\arabic{table}}
\setcounter{figure}{0} 
\setcounter{table}{0}

\section{Proof}
\label{app:proof}
\subsection{Proof of Theorem~\ref{thm:embedding_indep}}

\begin{proof}
By definition, the embedding of token $t$ is given by
\[
\widetilde{z}_{t} = E[t],
\]
which depends solely on the token identity $t$ and does not involve the positional index $p_t$. 

Formally, for any measurable function $f$ of the embedding,
\[
\mathbb{P}\big(f(\widetilde{z}_t) \leq x \,\big|\, p_t \big) 
= \mathbb{P}\big(f(\widetilde{z}_t) \leq x\big),
\]
because the distribution of $\widetilde{z}_t$ is completely determined by $t$ and does not involve $p_t$. 

Hence, the embedding $\widetilde{z}_t$ is independent of its positional index $p_t$.
\end{proof}

\subsection{Proof of Theorem~\ref{thm:unbiased}}

\begin{proof}
Let $\mu = \mathbb{E}[\widetilde{z}_{i,t}^{(j)}]$ denote the mean embedding vector. 
The aggregation is defined as
\[
\bar{z}_i^{(j)} = \frac{1}{n_i} \sum_{t=1}^{n_i} 
\frac{\mathrm{Concat}\big(p_t \widetilde{z}_{i,t}^{(j)},\ (l_{\max}-p_t) \widetilde{z}_{i,t}^{(j)}\big)}{l_{\max}}.
\]

Since the aggregation is linear in $\widetilde{z}_{i,t}^{(j)}$ and by Theorem~\ref{thm:embedding_indep} 
the embeddings are independent of $p_t$, taking expectation gives
\[
\mathbb{E}[\bar{z}_i^{(j)}] = \frac{1}{n_i l_{\max}} \sum_{t=1}^{n_i} 
\mathrm{Concat}\big(p_t \mu,\ (l_{\max}-p_t) \mu\big).
\]

Here, the expected aggregated vector depends only on the positional indices $p_t$ through the linear coefficients, 
and not on the actual embedding values. 
Therefore, each position contributes proportionally in expectation, and no single position $p_t$ 
receives undue influence. This establishes that the aggregation is statistically unbiased with respect to position.
\end{proof}

\section{Appendix For Method Details}
\label{app:method}
\subsection{Algorithm Pseudocode}
We present the pseudocode of ELILLM in Algorithm~\ref{alg:elillm}.
\begin{algorithm*}[tb]
\caption{Procedure of ELILLM}
\label{alg:elillm}
\textbf{Input}: Observed dataset \(\mathcal{D}_{\text{obs}} = \{(s_i, y_i)\}_{i=1}^{n}\), prompt \(p\), protein pocket \(t\), generation count \(k\)\\
\textbf{Output}: Generated molecule set \(\mathcal{S}_{\text{out}}\)
\begin{algorithmic}[1]
\STATE Initialize \(\mathcal{S}_{\text{out}} \leftarrow \emptyset\)
\WHILE{\(|\mathcal{S}_{\text{out}}| < k\)}
    \STATE Update latent observations \(\mathcal{Z}_{\text{obs}}\) using Equation~\ref{eq:zobs} \COMMENT{Project \(\mathcal{D}_{\text{obs}}\) into latent embedding space}
    \STATE Train surrogate model \(\hat{f}_\theta\) on \(\{(z_i, y_i)\}\) using Equations~\ref{eq:mse} and~\ref{eq:nll} \COMMENT{Fit surrogate model based on latent embeddings}
    \STATE Sample exploration embedding set \(\mathcal{Z}_{\text{explore}}\) using Equations~\ref{eq:perturb} and~\ref{eq:zexplore} \COMMENT{Generate candidate embeddings in latent space}
    \STATE Simulate evaluation of \(\mathcal{Z}_{\text{explore}}\) using surrogate model with Equation~\ref{eq:predict} \COMMENT{Simulated assessment via surrogate model}
    \STATE Select candidate embedding set \(\mathcal{Z}_{\text{cand}}\) based on Equation~\ref{eq:zcand} \COMMENT{Choose promising embeddings for generation}
    \STATE Decode candidates to molecules using Equation~\ref{eq:scand} \COMMENT{Generate candidate molecules from embeddings}
    \STATE Check validity and remove duplicates in \(\mathcal{S}_{\mathrm{cand}}\) \COMMENT{Filter invalid or repeated molecules}
    \STATE Evaluate filtered \(\mathcal{S}_{\mathrm{cand}}\) with docking score \(y = \text{Dock}(s, t)\) \COMMENT{Compute true docking scores using molecule \(s\) and protein pocket \(t\)}
    \STATE Update \(\mathcal{S}_{\text{out}} \leftarrow \mathcal{S}_{\text{out}} \cup \mathcal{S}_{\mathrm{cand}}\) \COMMENT{Add new valid molecules to output set}
    \STATE Update \(\mathcal{D}_{\text{obs}} \leftarrow \mathcal{D}_{\text{obs}} \cup \{(s, y) \mid s \in \mathcal{S}_{\mathrm{cand}}, y = \text{Dock}(s, t)\}\) \COMMENT{Augment observed data with new molecules and docking scores}
\ENDWHILE
\STATE \textbf{return} \(\mathcal{S}_{\text{out}}\)
\end{algorithmic}
\end{algorithm*}

\subsection{GP Kernel Details}
In our surrogate model \( \hat{f}_\theta \), we adopt a composite kernel that combines the Matérn-1.5 and Matérn-2.5 kernels~\cite{rf:matern} to better capture both local variations and global smoothness in the latent space. The resulting kernel is defined as:
\[
k(\mathbf{z}, \mathbf{z}') = \lambda_1 \cdot k_{\mathrm{Matern}\text{-}1.5}(\mathbf{z}, \mathbf{z}') + \lambda_2 \cdot k_{\mathrm{Matern}\text{-}2.5}(\mathbf{z}, \mathbf{z}'),
\]
where \( \lambda_1 \) and \( \lambda_2 \) are learnable weights.

The Matérn-1.5 kernel is given by
\[
k_{\mathrm{Matern}\text{-}1.5}(\mathbf{z}, \mathbf{z}') = \sigma^2 \left(1 + \frac{\sqrt{3} \|\mathbf{z} - \mathbf{z}'\|}{\ell} \right) \exp\left( - \frac{\sqrt{3} \|\mathbf{z} - \mathbf{z}'\|}{\ell} \right),
\]
and the Matérn-2.5 kernel is
\begin{align*}
k_{\mathrm{Matern}\text{-}2.5}(\mathbf{z}, \mathbf{z}') =\; & \sigma^2 \Bigg( 1 + \frac{\sqrt{5} \|\mathbf{z} - \mathbf{z}'\|}{\ell} \\
& + \frac{5 \|\mathbf{z} - \mathbf{z}'\|^2}{3\ell^2} \Bigg) 
\exp\left( - \frac{\sqrt{5} \|\mathbf{z} - \mathbf{z}'\|}{\ell} \right)
\end{align*}
where \( \ell \) is the length-scale and \( \sigma^2 \) is the output variance, both of which are learnable hyperparameters. The combination of Matérn-1.5 and Matérn-2.5 allows the model to benefit from the flexibility of the former in capturing sharp local transitions and the smoothness of the latter in modeling broader trends.
\subsection{Prompt}
We present the prompts used in our study in Figures~\ref{fig:elillm_prompt}--\ref{fig:no_role_prompt}: Figure~\ref{fig:elillm_prompt} shows the prompt designed for Knowledge-Guided LLM Decoding, Figure~\ref{fig:no_knowledge_prompt} shows the prompt used in the w/o knowledge ablation study, and Figure~\ref{fig:no_role_prompt} shows the prompt used in the w/o role ablation setting.

\begin{figure*}[tb]
\centering
\begin{tcolorbox}[mypromptbox,title=ELILLM Prompt]
Instructions:
You are a SMILES repair engine that must always generate the closest valid SMILES approximation for any input, never returning an empty string. You may use the following rules as appropriate, including but not limited to:  \\
Basic Repairs: \\
Complete missing ring numbers: C1CC → C1CC1\\
Balance parentheses: C(=O)(O → C(=O)O\\
Fix atom formatting: na → [Na]\\
Error Tolerance:\\
Illegal characters → replace with carbon or other atoms that you think are most likely: [X]C → CC\\
Uncloseable rings → convert to chains: C1CC2 → CCC\\
Hypervalent atoms → remove excess bonds: C(C)(C)(C)(C) → C(C)(C)C\\
Final Safeguard:\\
When completely unrecognizable:
Keep all valid atoms and single bonds: A*B → CB\\
Minimal output: /.\ → C\\
Examples:\\
(ring closure)Input: C1CC → Output: C1CC1\\
(invalid atom → C)Input: [X][Na+] → Output: C[Na+]\\
(remove unparseable stereochemistry)Input: C(/N)=C/F → Output: C(N)=CF\\
(wildcard → C)Input: *C(=O)O → Output: CC(=O)O
(keep only valid atoms)Input: A1B2 → Output: CC\\
Output Specification:\\
Only return the repaired SMILES string, nothing else. No explanations, no confirmations, just the valid SMILES. If input is already valid, return it unchanged.
\end{tcolorbox}
\caption{ELILLM prompt}
\label{fig:elillm_prompt}
\end{figure*}

\begin{figure*}[tb]
\centering
\begin{tcolorbox}[mypromptbox,title=w/o knowledge Prompt]
Output a SMILES string only!\\
You are a SMILES repair engine that must always generate the closest valid SMILES approximation for any input, never returning an empty string.\\
If the input string is already a valid SMILES, output this SMILES only!!
\end{tcolorbox}
\caption{w/o knowledge prompt}
\label{fig:no_knowledge_prompt}
\end{figure*}

\begin{figure*}[tb]
\centering
\begin{tcolorbox}[mypromptbox,title=w/o role Prompt]
Output a SMILES string only!\\
Input a SMILES string to you, and your task is to output this SMILES string.
\end{tcolorbox}
\caption{w/o role prompt}
\label{fig:no_role_prompt}
\end{figure*}

\begin{table*}[htbp]
\centering
\caption{One-sided Wilcoxon signed-rank test $p$-values assessing the statistical significance that ELILLM-diff achieves significantly lower docking scores compared to some baseline methods across Top-K results}
\label{tab:docking_pvalues}
\begin{tabular}{lcccc}
\toprule
\textbf{Methods} 
& \textbf{Top1 Dock $p$}
& \textbf{Top5 Dock $p$} 
& \textbf{Top10 Dock $p$}  
& \textbf{Top20 Dock $p$} \\
\midrule
LiGAN & 1.33e-14 & 5.70e-15 & 5.40e-15 & 1.42e-14 \\
Pocket2Mol & 2.71e-5 & 1.70e-4 & 4.37e-4 & 1.45e-3\\
TargetDiff & 7.72e-4 & 1.21e-6 & 3.85e-8 & 1.01e-9 \\
ALIDIFF & 1.13e-5 & 1.19e-9 & 3.41e-12 & 6.50e-14 \\

\bottomrule
\end{tabular}
\end{table*}

\section{Detailed Experimental Settings}
\label{app:expsetting}
\subsection{Hyperparameter Settings}  
For our method, we explored a series of hyperparameter ranges across architecture design, surrogate model training, and generation-time settings. 

In terms of \textbf{architecture}, we adopted an MLP structure of 8192-256-256-256-20, corresponding to three hidden layers. The hidden layer size was searched over $\{128, 256, 512\}$, and 256 was selected as the final choice. The latent dimension was chosen from $\{10, 15, 20, 50\}$, with 20 providing the best performance. For the Gaussian Process (GP) component, we tested several kernel types including RBF, linear, Matérn, and their combinations, and finally selected a hybrid of Matérn-1.5 and Matérn-2.5 kernels.
For \textbf{surrogate model training}, the MLP was trained for 100 epochs in ELILLM-rand and 200 epochs in ELILLM-diff, while the GP model was trained for 100 epochs. The learning rate for the MLP was searched over $\{0.001, 0.005, 0.01\}$, and 0.001 was chosen. For the GP, the learning rate was tuned within $\{0.05, 0.1\}$, and 0.1 was selected. The Adam optimizer was used throughout training.
Regarding \textbf{generation-time parameters}, the perturbation scale $\lambda_{\mathrm{perturb}}$ was tuned over $\{0.3, 0.4, 0.5\}$, with 0.4 adopted as the final value. The maximum token length $l_{\mathrm{max}}$ was selected from $\{60, 80, 100\}$, and set to 80. The number of candidates $n_{\mathrm{candidate}}$ was fixed at 5. The sampling temperature of the LLM was tuned over $\{0.4, 0.5\}$, with 0.4 selected, and the acquisition function threshold $\delta$ was set to 0.1. All LLM-based generation was performed using a locally deployed LLaMA 3.1 8B model.

For the baselines, we followed the default hyperparameters provided in their original papers or official implementations. To ensure a fair comparison, we re-implemented LMLF based on LLama 3.1 8B. To guarantee that LMLF can run successfully on the task of generating 100 ligands, we set the \textit{docking threshold} to $-6.5$, and the \textit{threshold increment frequency} to 50, meaning the threshold decreases by 1 every 50 iterations. In line with our setup, the number of ligands generated per round was capped at 5. All generated molecules, including those from baselines, were evaluated using our docking pipeline to ensure fair comparison. Specifically, we used RDKit to parse the generated SMILES strings, construct initial 3D coordinates via molecular embedding, and optimize them using the MMFF94 force field. Vina docking scores were then computed using Smina with the default scoring function (vina). The exhaustiveness was set to 32, the number of poses was kept at the default value of 9, and the binding pocket was determined based on the known co-crystallized ligand of the target protein. The autobox-add parameter was set to 1 Å to define the docking box around the pocket. Among all generated poses, we selected the one with the lowest docking score as the final Vina score.

\subsection{System and Software Configuration}

The experiments under the \textbf{ELILLM-rand} and \textbf{ELILLM-diff} settings were conducted on two separate machines, both equipped with \textbf{NVIDIA RTX 3090 GPUs}. 

For the \textbf{ELILLM-rand} setting, the system was configured with an Intel(R) Xeon(R) CPU E5-2673 v4 @ 2.30GHz and ran on Ubuntu 20.04.6 LTS. For the \textbf{ELILLM-diff} setting, the hardware setup included an Intel(R) Xeon(R) Silver 4310 CPU @ 2.10GHz, also running Ubuntu 20.04.6 LTS. 

Both experiments shared the same major software environment, including Python 3.10, PyTorch 2.3.1, HuggingFace Transformers 4.51.3, RDKit 2024.9.6, Smina 2020.12.10, and CUDA 12.1.

\subsection{Code and Reproducibility}
To ensure reproducibility, we include all code and scripts necessary for training, evaluation, and visualization in an anonymous online repository at https://github.com/hxnhxn/ELILLM. All experiments are conducted under controlled settings with fixed random seeds (e.g., seed = 1) to minimize randomness and ensure consistency.

\section{Supplementary Experiment}
\label{app:addexp}
\subsection{One-sided Wilcoxon Signed-Rank Test}
To assess the statistical significance of our method’s performance improvement, we conduct one-sided Wilcoxon signed-rank~\cite{rf:wilcoxon} tests on the average docking scores (lower is better) across 100 target proteins under Top-1, Top-5, Top-10, and Top-20 candidate settings. Specifically, we compare our method against several strong baseline methods with the best reported performance.

The null hypothesis for each test is that our method does not achieve significantly lower docking scores than the compared baseline (i.e., the paired differences are symmetrically distributed around zero or skewed in the opposite direction). The alternative hypothesis is that our method achieves significantly lower scores, indicating superior binding affinity.

The test is non-parametric and appropriate for our setting where paired scores are compared across the same set of protein targets, but normality of differences cannot be assumed. Reported p-values reflect the confidence with which we can reject the null hypothesis and support the superiority of our method. As shown in Table~\ref{tab:docking_pvalues}, the $p$-values in all settings are below the 0.05 significance threshold, indicating that our method is significantly better than the selected baselines across all Top-$K$ docking score evaluations.
\subsection{Additional Visualization}
\begin{figure}[h]
    \centering
    \includegraphics[width=1\linewidth]{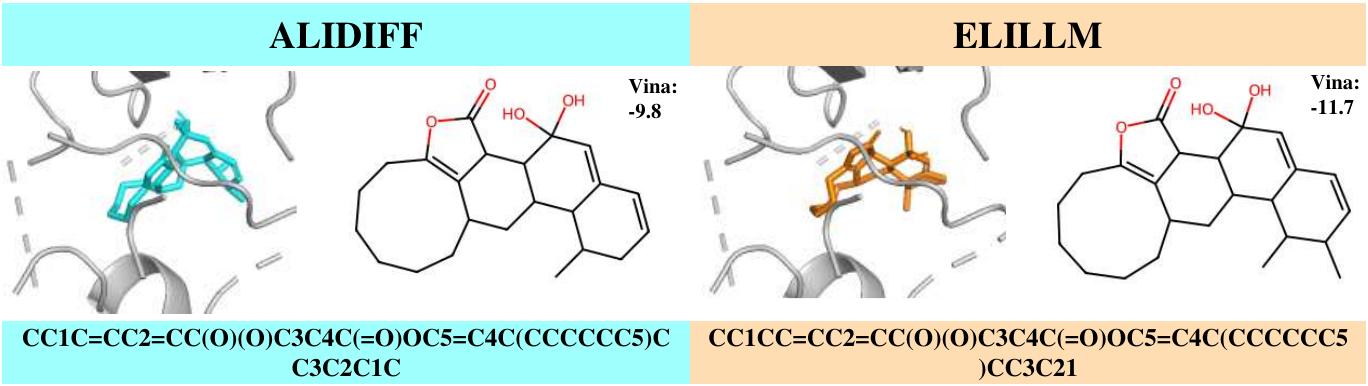}
    \caption{Visualizations of generated ligands for protein pockets 3nfb generated by ALIDIFF and ELILLM-diff.}
    \label{fig:add_visualization}
\end{figure}
While the visualizations in the main paper highlight the exploration capability of ELILLM, its ability to exploit existing molecular structures is equally important. As shown in Figure~\ref{fig:add_visualization}, ELILLM achieves substantial improvements in binding affinity by making slight modifications to the original structure while preserving its overall scaffold.
\end{document}